# Convergence Rates for Differentially Private Statistical Estimation

**Kamalika Chaudhuri**      KAMALIKA@CS.UCSD.EDU
University of California, San Diego, La Jolla, CA 92093

**Daniel Hsu**      DAHSU@MICROSOFT.COM
Microsoft Research, New England, Cambridge, MA 02142

## Abstract

Differential privacy is a cryptographically-motivated definition of privacy which has gained significant attention over the past few years. Differentially private solutions enforce privacy by adding random noise to a function computed over the data, and the challenge in designing such algorithms is to control the added noise in order to optimize the privacy-accuracy-sample size tradeoff.

This work studies differentially-private statistical estimation, and shows upper and lower bounds on the convergence rates of differentially private approximations to statistical estimators. Our results reveal a formal connection between differential privacy and the notion of Gross Error Sensitivity (GES) in robust statistics, by showing that the convergence rate of any differentially private approximation to an estimator that is accurate over a large class of distributions has to grow with the GES of the estimator. We then provide an upper bound on the convergence rate of a differentially private approximation to an estimator with bounded range and bounded GES. We show that the bounded range condition is necessary if we wish to ensure a strict form of differential privacy.

## 1. Introduction

Differential privacy (Dwork et al., 2006b) is a strong, cryptographically-motivated definition of privacy which has gained significant attention in the machine-learning and data-mining communities over the past few years (McSherry & Mironov, 2009; Chaudhuri et al., 2011; Friedman & Schuster, 2010; Mohammed et al., 2011). In differentially private solutions, privacy is guaranteed by ensuring that the participation of a single individual in a database does not change the outcome of a private algorithm by much. This is typically achieved by adding some random noise, either to the sensitive input data, or to the output of some function, such as a classifier, computed on the sensitive data. While this guarantees privacy, for most statistical and machine learning tasks, there is a subsequent loss in *statistical efficiency*, in terms of the number of samples required to estimate a function to a given degree of accuracy. Thus the main challenge in designing differentially private algorithms is to optimize the privacy-accuracy-sample size trade-off, and a body of literature has been devoted to this goal.

In this paper, we focus on differentially-private statistical estimation. We ask: what properties should a statistical estimator have, so that it can be approximated accurately with differential privacy? Privately approximating an estimator based on a functional $T$ that performs well when data is drawn from a *specific* distribution $F$ is easy: ignore the sensitive data, and output $T(F)$. Thus the challenge is to design differentially private approximations to estimators that are accurate over a wide range of distributions.

Previous work (Smith, 2011) on differentially private statistical estimation shows how to construct differentially private approximations to estimators which have asymptotic normality guarantees under fairly mild conditions. In practical situations, however, we must take into account the effect of a finite number of samples. Moreover, it has been empirically observed (*e.g.*, Chaudhuri et al., 2011; Vu & Slavkovic, 2009) that there is often a significant gap in statistical efficiency between a differentially private estimator and its non-private counterpart. Thus there is a need to study finite sample convergence rates for differentially private statistical estimators, in order to characterize the properties that make a statistical estimator amenable to differentially-private approximations.

In this paper, we provide upper and lower bounds on the finite sample convergence rates of such estimators. Our first finite sample result draws a connection between differentially private statistical estimators and *Gross Error Sensitivity*, a measure commonly used in the robust statistics literature (Huber, 1981). The Gross Error Sensitivity (GES) of a statistical functional $T$ at a distribution $F$ is the maximum change in the value of $T(F)$ by an arbitrarily small perturbation of $F$ by any point mass $x$ in the domain. We provide a lower bound on the convergence rate of any differentially private statistical estimator, showing that an estimator that approximates $T(F_n)$ well with differential privacy over a large class of distributions must have its convergence rate grow with the GES of $T$.

A natural question to ask next is whether bounded GES is sufficient for the existence of differentially private estimators that are accurate for large classes of distributions. We next show that at least for $\alpha$-differential privacy, this is not the case. Any estimator based on a functional $T$ that takes values in a range of length $R$ and guarantees $\alpha$-differential privacy for a wide class of distributions, has to have a finite sample convergence rate that grows with increasing $R$.

We then show that bounded range and GES are indeed sufficient for differentially private estimation. In particular, given an estimator based on a functional $T$ which takes values in a bounded range, and has bounded GES for all distributions *close* to the underlying data distribution $F$, we show how to compute a differentially private approximation to $T(F)$ based on sensitive data drawn from $F$. Our approximation preserves $(\alpha, \delta)$-differential privacy, a relaxation of $\alpha$-differential privacy, and is based on the smoothed sensitivity method (Nissim et al., 2007). We provide a finite sample upper bound on the convergence rate of this estimator.

The statistical estimators in our upper bounds are computationally inefficient in general. We conclude by providing a separate explicit method for privately approximating M-estimators with certain properties. We prove that these differentially-private estimators enjoy similar privacy and statistical guarantees as those based on the smooth-sensitivity method, while being more efficiently computable.

**Related Work**

Differential privacy was proposed by (Dwork et al., 2006b), and has been used since in many works on privacy (*e.g.*, Blum et al., 2005; Barak et al., 2007; Nissim et al., 2007; McSherry & Mironov, 2009; Chaudhuri et al., 2011). It has been shown to have strong semantic guarantees (Dwork et al., 2006b) and is resistant to many attacks (Ganta et al., 2008) that succeed against some other definitions of privacy.

Dwork & Lei (2009) is the first work to identify a connection between differential privacy and robust statistics; based on robust statistical estimators as a starting point, they provide differentially private algorithms for several common estimation tasks, including interquartile range, trimmed mean and median, and regression.

In further work, Smith (2011) shows how to construct a differentially private approximation $\mathcal{A}_T$ to certain types of statistical estimators $T$, and establishes asymptotic normality of his estimator provided certain conditions on $T$ hold. We in contrast focus on finite sample bounds, with an aim towards characterizing the statistical properties of estimators that determine how closely they can be approximated with differential privacy. Lei (2011) considers M-estimation, and provides a simple and elegant differentially-private M-estimator which is statistically consistent.

Finally, work on the sample requirement of differentially private algorithms include bounds on the accuracy of differentially private data release (Hardt & Talwar, 2010), and the sample complexity of differentially private classification (Chaudhuri & Hsu, 2011).

## 2. Preliminaries

The goal of this paper is to examine the conditions under which we can find private approximations to estimators. The notion of privacy we use is differential privacy (Dwork et al., 2006b;a).

**Definition 1.** A (randomized) algorithm $\mathcal{A}$ taking values in a range $\mathcal{S}$ is $(\alpha, \delta)$-*differentially private* if for all $S \subseteq \mathcal{S}$, and all data sets $D$ and $D'$ differing in a single entry,

$$\Pr_{\mathcal{A}}[\mathcal{A}(D) \in S] \leq e^\alpha \Pr_{\mathcal{A}}[\mathcal{A}(D') \in S] + \delta,$$

where $\Pr_{\mathcal{A}}[\cdot]$ is the distribution on $\mathcal{S}$ induced by the output of $\mathcal{A}$ given a data set.

A (randomized) algorithm $\mathcal{A}$ is $\alpha$-*differentially private* if it is $(\alpha, 0)$-differentially private.

Here $\alpha > 0$ and $\delta \in [0, 1]$ are privacy parameters, where smaller $\alpha$ and $\delta$ imply stricter privacy.

A general approach to developing differentially private approximations to functions is to add noise, either to the sensitive data, or to the output of a non-private function computed on the data. This work explores what properties statistical functionals need to have so that they can be accurately approximated with differential privacy.

Let $\mathcal{F}$ denote the space of probability distributions on a domain $\mathcal{X}$. A *statistical functional* $T\colon \mathcal{F} \to \mathbb{R}$ is a real-valued function of a distribution $F$. The *plug-in estimator* of $\theta = T(F)$ is given by $\theta_n := T(F_n)$, where $F_n$ is the empirical distribution corresponding to an i.i.d. sample of size $n$ drawn from $F$.

A common measure of the robustness of a statistical functional is the *influence function*, which measures how a functional $T(F)$ responds to small changes to the input $F$.

**Definition 2.** The *influence function* $\mathbf{IF}(x, T, F)$ for a functional $T$ and distribution $F$ at $x \in \mathcal{X}$ is:

$$\mathbf{IF}(x, T, F) = \lim_{\rho \to 0} \frac{T((1-\rho)F + \rho \delta_x) - T(F)}{\rho}$$

where $\delta_x$ denotes the point mass distribution at $x$.

It is a well-established result in theoretical statistics (see, *e.g*, Wasserman, 2006) that if $T$ is Hadamard-differentiable, and if $\mathbb{E}_{x \sim F}[\mathbf{IF}(x, T, F)^2]$ is bounded, then $T(F_n)$ converges to $T(F)$ as $n \to \infty$.

A related notion is that of *gross error sensitivity*, which measures the worst-case value of the influence function for any $x \in \mathcal{X}$.

**Definition 3.** The *gross error sensitivity* $\mathbf{GES}(T, F)$ for a functional $T$ and distribution $F$ is:

$$\mathbf{GES}(T, F) = \sup_{x \in \mathcal{X}} |\mathbf{IF}(x, T, F)|.$$

We also define the notions of influence function and gross error sensitivity at a fixed scale $\rho > 0$:

$$\mathbf{IF}_\rho(x, T, F) := \frac{T((1-\rho)F + \rho \delta_x) - T(F)}{\rho}$$
$$\mathbf{GES}_\rho(T, F) := \sup_{x \in \mathcal{X}} |\mathbf{IF}_\rho(x, T, F)|.$$

In this work, the data domain $\mathcal{X}$ will be a subset of $\mathbb{R}$. We overload notation and use $F$ to denote a distribution as well as its cumulative distribution function. For two distributions $F$ and $G$, we use $d_{\mathrm{GC}}(F, G) := \sup_{x \in \mathbb{R}} |F(x) - G(x)|$ to denote the Glivenko-Cantelli distance between $F$ and $G$. For a distribution $F$ from a family $\mathcal{F}$ and a radius $r > 0$, let $\mathcal{B}_{\mathrm{GC}}(F, r)$ denote the set of distributions $G \in \mathcal{F}$ such that $d_{\mathrm{GC}}(F, G) \leq r$. Finally, we use $d_{\mathrm{TV}}(F, G)$ to denote the total variation distance between $F$ and $G$.

A statistical functional $T$ is *B-robust* at $F$ if $\mathbf{GES}(T, F)$ is finite. B-robustness has been studied in the robust statistics literature (Hampel et al., 1986; Huber, 1981), and plug-in estimators for B-robust functionals are considered to be resistant to outliers and changes in the input.

## 3. Lower Bounds

We begin by establishing lower bounds on the convergence rate of any differentially private approximation to a statistical functional $T(F)$.

### 3.1. Lower Bounds based on Gross Error Sensitivity

We first show a lower bound on the error of any $(\alpha, \delta)$-differentially private approximation to $T$ in terms of the gross error sensitivity of $T$ at a distribution $F$.

**Theorem 1.** *Pick any $\alpha \in (0, \frac{\ln 2}{2})$ and $\delta \in (0, \frac{\alpha}{23})$. Let $\mathcal{F}$ be the family of all distributions over $\mathcal{X}$, and let $\mathcal{A}$ be any $(\alpha, \delta)$-differentially private algorithm. For all $n \in \mathbb{N}$ and all $F \in \mathcal{F}$, there exists a radius $\rho = \rho(n) = \frac{1}{n} \cdot \lceil \frac{\ln 2}{2\alpha} \rceil$ and a distribution $G \in \mathcal{F}$ with $d_{\mathrm{TV}}(F, G) \leq \rho$, such that either*

$$\mathbb{E}_{F_n \sim F} \mathbb{E}_\mathcal{A}[|\mathcal{A}(F_n) - T(F)|] \geq \frac{\rho}{16} \mathbf{GES}_\rho(T, F), \text{ or}$$

$$\mathbb{E}_{G_n \sim G} \mathbb{E}_\mathcal{A}[|\mathcal{A}(G_n) - T(G)|] \geq \frac{\rho}{16} \mathbf{GES}_\rho(T, F).$$

Several remarks are in order. First of all, the form of Theorem 1 is slightly unconventional in the sense that applies not to particular distributions, but to a set of distributions. In particular, the bound states that either the convergence rate of $F$ is high, or the convergence rate of some $G$ close to $F$ is high. Observe that for a fixed distribution $F$, it is trivial to construct a differentially private approximation to $T(F)$ that is accurate for $F$ – ignore any sensitive input data, and simply output $T(F)$. This algorithm provides a perfectly accurate estimate when the input is drawn from $F$, but performs poorly otherwise; thus any lower bound that applies to all differentially private algorithms will have a similar form. On the other hand, the differentially private estimators in Theorem 1 have few restrictions: they are only expected to be accurate for distributions lying in a small neighborhood of $F$, and may be extremely inaccurate in general.

Second, for fixed $n$, $\rho$ is a function $\rho(n) = \frac{1}{n} \cdot \lceil \frac{\ln 2}{2\alpha} \rceil$, which decreases to zero as $n \to \infty$; provided $\mathbf{GES}_\rho(T, F)$ remains the same as $\rho$ diminishes, the lower bound grows weaker with increasing $n$. The lower bound thus does not rule out the existence of consistent private estimators.

Finally, we observe from the proof of Theorem 1 that $\mathcal{F}$ need not be the family of all distributions over $\mathcal{X}$; the theorem will still apply if for every $F \in \mathcal{F}$, and for all $x \in \mathcal{X}$, $(1-\rho)F + \rho\delta_x$ also lies in the family $\mathcal{F}$; for example if $\mathcal{F}$ is the set of all discrete distributions over $\mathcal{X}$.

While Theorem 1 is very general, we present below an example that illustrates an implication of the theorem.

**Example 1.** Let $\mathcal{X} = [0, a]$, and let $\mathcal{F}$ be the set of all discrete distributions over $\mathcal{X}$. Let $T(F)$ be the mean of $F$.

Cosnider a fixed $F \in \mathcal{F}$, and a fixed $n$. Let $\rho = \rho(n)$ as in Theorem 1. For any $F$, $\mathbf{GES}_\rho(T, F) \geq \frac{a}{2}$. It can be shown that for any $G \in \mathcal{B}_{\mathrm{TV}}(F, \rho(n))$, $\mathrm{Var}[G] \leq \mathrm{Var}[F] + \rho(1-\rho)a^2$. Thus, the expected errors of the (non-private) plug-in estimators are bounded as $\mathbb{E}[|T(F_n) - T(F)|] \leq O(\sqrt{\mathrm{Var}[F]/n})$ and $\mathbb{E}[|T(G_n) - T(G)|] \leq O(\sqrt{\mathrm{Var}[F]/n} + \sqrt{\rho(1-\rho)a^2/n})$ for all $G \in \mathcal{B}_{\mathrm{TV}}(F, \rho(n))$. On the other hand, Theorem 1 shows that for every differentially private estimator $\mathcal{A}$, at least one of $\mathbb{E}[|\mathcal{A}(F_n) - T(F)|]$ and $\mathbb{E}[|\mathcal{A}(G_n) - T(G)|]$ is $\Omega(\rho a)$; this quantity is higher than the corresponding quantity for the non-private estimator so long as $n \leq O(\frac{a^2}{\mathrm{Var}[F]\alpha^2})$.

*Proof of Theorem 1.* Let $x^*$ be the $x \in \mathcal{X}$ that maximizes $|\mathbf{IF}_\rho(x, T, F)|$. Let $\gamma > 0$, and let $\rho := \frac{1}{n}\lceil \frac{\ln 2}{2\alpha} \rceil$, and let $G := (1-\rho)F + \rho\delta_{x^*}$. Observe that $d_{\mathrm{TV}}(F, G) \leq \rho$ and $\mathbf{IF}_\rho(x^*, T, F) = (T(G) - T(F))/\rho$.

Consider the following procedure for drawing $n$ samples from $G$. First, draw a random sample $F_n$ of size $n$ from $F$ (we overload the notation $F_n$ to refer to both a random sample and its empirical distribution). Next, for each $i = 1, 2, \ldots, n$, independently toss a biased coin with heads probability $\rho$; if the coin turns up heads, replace the $i$-th element of $F_n$ by $x^*$; otherwise, do nothing. This procedure constructs a random sample $G_n$ of size $n$ from $G$, and in the process constructs a coupling between samples of size $n$ from $F$ and $G$. In what follows, we will use this coupling to calculate the quantity

$$\mathbb{E}_{F_n \sim F}\mathbb{E}_{\mathcal{A}}[|\mathcal{A}(F_n) - T(F)|] + \mathbb{E}_{G_n \sim G}\mathbb{E}_{\mathcal{A}}[|\mathcal{A}(G_n) - T(G)|].$$

Let $F_n$ be any randomly drawn sample of size $n$ from $F$, and let $G_n$ be a corresponding sample from $G$ as drawn from the coupling procedure. Call a pair $(F_n, G_n)$ $\rho$-close if they differ in at most $\rho n$ entries. As the median of $\mathrm{Binomial}(n, \rho)$ is $\leq \lceil \rho n \rceil = \rho n$, the probability that at most $\rho n$ of the elements of $F_n$ are converted to $x^*$ by the coupling process is at least $1/2$.

In other words,

$$\Pr_{G_n}[(F_n, G_n) \text{ is } \rho\text{-close}] \geq 1/2. \quad (1)$$

For any $\rho$-close pair $(F_n, G_n)$, we can apply Lemma 3[1] with the parameters $t := T(F)$, $t' := T(G)$, $\gamma := 1/4$, and

$$\Delta := \rho n \leq \left(1 + \frac{\ln 2}{2\alpha}\right) \leq \frac{\ln 2}{\alpha} = \frac{\ln \frac{1}{2\gamma}}{\alpha};$$

the lemma implies, for any $\rho$-close pair $(F_n, G_n)$,

$$\mathbb{E}_{\mathcal{A}}[|\mathcal{A}(F_n) - T(F)|] + \mathbb{E}_{\mathcal{A}}[|\mathcal{A}(G_n) - T(G)|]$$
$$\geq \frac{1}{4}|T(F) - T(G)|.$$

Therefore, conditioned on $F_n$, we have

$$\mathbb{E}_{\mathcal{A}}[|\mathcal{A}(F_n) - T(F)|] + \mathbb{E}_{G_n}\mathbb{E}_{\mathcal{A}}[|\mathcal{A}(G_n) - T(G)||F_n]$$
$$\geq \frac{1}{8}|T(F) - T(G)|$$

by (1). Taking a final expectation over $F_n \sim F$,

$$\mathbb{E}_{F_n \sim F}\mathbb{E}_{\mathcal{A}}[|\mathcal{A}(F_n) - T(F)|] + \mathbb{E}_{G_n \sim G}\mathbb{E}_{\mathcal{A}}[|\mathcal{A}(G_n) - T(G)|]$$
$$\geq \frac{1}{8}|T(F) - T(G)|$$
$$= \frac{\rho}{8}|\mathbf{IF}_\rho(x^*, T, F)| = \frac{\rho}{8}\mathbf{GES}_\rho(T, F).$$

The theorem follows. $\square$

### 3.2. Lower Bounds as a Function of Range

Is the bound in Theorem 1 tight? In other words, if $T$ has bounded GES, can we compute accurate differentially private approximations to $T(F)$ for all distributions $F$ over a domain? We next show that at least for $(\alpha, 0)$-differential privacy, Theorem 1 is not tight; if we wish to compute differentially private and accurate estimates of $T(F)$ for all distributons $F$ in a family, where $T(F)$ can take any value in a range $[\lambda, \lambda']$, then the sample size must grow as a function of $\lambda' - \lambda$.

**Theorem 2.** *Let $\mathcal{F}$ be a family of distributions over $\mathcal{X}$, and let $\mathcal{A}$ be any $(\alpha, 0)$-differentially private algorithm. Suppose for all $\tau \in [\lambda, \lambda']$, there exists some $F^\tau \in \mathcal{F}$ such that $T(F^\tau) = \tau$. Then there exists some $F \in \mathcal{F}$ such that*

$$\mathbb{E}_{F_n \sim F, \mathcal{A}}[|\mathcal{A}(F_n) - T(F)|] \geq \frac{1}{4} \cdot \frac{\lambda' - \lambda}{2 + e^{\alpha n}}.$$

**Example 2.** For any $\gamma \in \mathbb{R}$, let $U_\gamma$ be the uniform distribution on $[\gamma - 1, \gamma + 1]$, and let $\mathcal{F}$ be the family $\mathcal{F} = \{U_\gamma : \gamma \in [-R, R]\}$. Let $T(F)$ be the median of $F$. For every $F \in \mathcal{F}$, the non-private estimator $T(F_n)$ converges to $T(F)$ at a rate proportional to $O(\frac{1}{\sqrt{n}})$, independent of $R$. However, Theorem 2 shows that for every differentially private estimator $\mathcal{A}$, there is some $F \in \mathcal{F}$ such that $|\mathcal{A}(F_n) - T(F)|$ grows with $R$.

---

[1] See Appendix A for omitted lemmas.

*Proof of Theorem 2.* Let $r := \frac{\lambda' - \lambda}{2 + e^{\alpha n}}$ and $\Gamma := \lfloor \frac{\lambda' - \lambda}{r} \rfloor$. For each $i = 1, 2, \ldots, \Gamma$, let $F^i$ be a distribution in $\mathcal{F}$ such that $T(F^i) = \lambda + (i - \frac{1}{2})r$; such distributions are guaranteed to exist by assumption. Also, for each $i = 1, 2, \ldots, \Gamma$, let $F_n^i$ be an iid sample of size $n$ from $F^i$, and define the half-open interval $I^i := [\lambda + (i - 1)r, \lambda + ir)$. Observe that the intervals $I_i$ are disjoint. To prove the theorem, let us assume the contrary:

$$\mathbb{E}_{F_n^i, \mathcal{A}}[|\mathcal{A}(F_n^i) - T(F^i)|] \leq r/4 \quad \text{for all } i. \quad (2)$$

This, along with a Markov's inequality on $|\mathcal{A}(F_n^i) - T(F^i)|$, implies that $\Pr_{F_n^i, \mathcal{A}}[\mathcal{A}(F_n^i) \in I^i] \geq 1/2$. Therefore, for any $i$,

$$\frac{1}{2} \geq \Pr_{F_n^i, \mathcal{A}}[\mathcal{A}(F_n^i) \notin I^i] \geq \sum_{j \neq i} \Pr_{F_n^i, \mathcal{A}}[\mathcal{A}(F_n^i) \in I^j]$$

$$\geq e^{-\alpha n} \sum_{j \neq i} \Pr_{F_n^j, \mathcal{A}}[\mathcal{A}(F_n^j) \in I^j] \geq \frac{1}{2}(\Gamma - 1)e^{-\alpha n}$$

where the first step follows by assumption, the second step follows because the intervals $\{I^j\}$ are disjoint, and the third step from Lemma 2 and the fact that for any $i$ and $j$, any $F_n^i$ and $F_n^j$ differ in at most $n$ entries. Rearranging, the inequality becomes $\Gamma \leq 1 + e^{\alpha n}$, which is a contradiction since $\Gamma = \lfloor (\lambda' - \lambda)/r \rfloor > 1 + e^{\alpha n}$. Therefore (2) cannot hold, so the theorem follows. □

## 4. Upper Bounds

In this section, we show that bounded GES and bounded range are sufficient conditions for the existence of an $(\alpha, \delta)$-differentially private approximation to $T$. Our approximation uses the smooth-sensitivity method of Nissim et al. (2007), for which we provide a new statistical analysis in Section 4.1 (Theorem 3). We also provide a specific analysis for the case of linear functionals in Appendix B.

Let $d_H(D, D')$ denote the Hamming distance between $D$ and $D'$ (the number of entries in which $D$ and $D'$ differ), and recall the following definitions from Nissim et al. (2007).

**Definition 4.** The *local sensitivity* of a function $\varphi \colon \mathbb{R}^n \to \mathbb{R}$ at a data set $D \in \mathbb{R}^n$, denoted by $\mathrm{LS}(\varphi, D)$, is

$$\mathrm{LS}(\varphi, D) := \sup\{|\varphi(D) - \varphi(D')| : d_H(D, D') = 1\}.$$

For $\beta > 0$, the $\beta$-*smooth sensitivity* of $\varphi$ at $D$, denoted by $\mathrm{SS}_\beta(\varphi, D)$, is

$$\mathrm{SS}_\beta(\varphi, D) := \sup\{e^{-\beta d_H(D, D')} \cdot \mathrm{LS}(\varphi, D') : D' \in \mathbb{R}^n\}.$$

Throughout, we assume $D \in \mathbb{R}^n$ is an i.i.d. sample of size $n$ drawn from a fixed distribution $F$, and $F_n$ is the empirical CDF corresponding to this sample. For a statistical functional $T$, we use the overloaded notation $\mathrm{SS}_\beta(T, F_n)$ to denote the $\beta$-smooth sensitivity of $T(F_n)$ at the data set $F_n = D$.

### 4.1. Estimator Based on Smooth Sensitivity

For a statistical functional $T$, let $\mathcal{A}_T$ be the randomized estimator given by

$$\mathcal{A}_T(F_n) := T(F_n) + \mathrm{SS}_{\beta(\alpha, \delta)}(T, F_n) \cdot \frac{2}{\alpha} \cdot Z \quad (3)$$

where $\beta(\alpha, \delta) := \frac{\alpha}{2 \ln(1/\delta)}$ and $Z$ is an independent random variable drawn from the standard Laplace density $p_Z(z) = 0.5 e^{-|z|}$. $\mathcal{A}_T$ essentially computes $T(F_n)$ and adds zero-mean noise, with the scale determined by the privacy parameters and the smooth sensitivity. Computing $\mathrm{SS}_{\beta(\alpha, \delta)}(T, F_n)$ in general can be computationally challenging –see Nissim et al. (2007); our result thus demonstrates an upper bound.

The following guarantee is due to Nissim et al. (2007).

**Proposition 1.** $\mathcal{A}_T$ *is* $(\alpha, \delta)$-*differentially private.*

To give a statistical guarantee for $\mathcal{A}_T$, we begin with a standard tail bound based on the simple fact that $\Pr_Z[|Z| > t] \leq e^{-t}$.

**Proposition 2.** *For any* $t > 0$,

$$\Pr_Z\left[|\mathcal{A}_T(F_n) - T(F_n)| > \mathrm{SS}_{\beta(\alpha, \delta)}(T, F_n) \cdot \frac{2}{\alpha} \cdot t\right] \leq e^{-t}.$$

It follows that the convergence rate of $\mathcal{A}_T$ depends on the $\beta$-smooth sensitivity of $T$ at $F_n$, which can be bounded under the following conditions on $T$.

**Condition 1** (Bounded range). There exists a finite $R > 0$ such that the range of $T$ is contained in an interval of length $R$.

**Condition 2** (Bounded gross error sensitivity). The sequence $(\Gamma_n)$ given by

$$\Gamma_n := \sup\left\{\mathbf{GES}_{1/n}(T, G) : G \in \mathcal{B}_{\mathrm{GC}}\left(F, \sqrt{\frac{2 \ln(2/\eta)}{n}}\right)\right\}$$

is bounded.

Even for non-private estimation, the robustness of an estimator depends not just on the influence functions at the target distribution $F$, but also on these quantities in a local neighborhood around $F$ (Huber, 1981, p. 72). For convenience, Condition 2 is stated in terms of Glivenko-Cantelli distance, but can be easily changed to any distance under which $F_n$ converges to $F$ as $n \to \infty$ with suitable modifications in the analysis.

We now state our main statistical guarantee for $\mathcal{A}_T$.

**Theorem 3.** *Assume Condition 1 and Condition 2 hold. Pick any $\eta \in (0, 1/4)$. With probability $\geq 1 - 2\eta$, the estimator $\mathcal{A}_T$ from (3) satisfies*

$$|\mathcal{A}_T(F_n) - T(F)| \leq |T(F_n) - T(F)| +$$
$$\frac{2\ln(1/\eta)}{\alpha} \max\left\{\frac{2\Gamma_n}{n}, R \cdot \exp\left(-\frac{\alpha\sqrt{n\ln(2/\eta)}}{74\ln(1/\delta)}\right)\right\}$$

*where $R$ is the quantity in Condition 1, and $\Gamma_n$ is the quantity in Condition 2.*

*Proof.* Follows from Proposition 2, Lemma 1, a union bound, and the triangle inequality. □

The first term in the bound, $|T(F_n) - T(F)|$, is the error of the non-private plug-in estimate $T(F_n)$. If $T$ is Hadamard-differentiable, then $T(F_n) - T(F)$ converges in distribution to a zero-mean normal random variable with variance $n^{-1} \int \mathbf{IF}(x, T, F)^2 dF(x)$; in this case, $T(F_n)$ converges to $T(F)$ at an asymptotic $n^{-1/2}$ rate (Wasserman, 2006). Non-asymptotic rates can also be established in terms of other specific properties of $T$ and $F$ (see Appendix B for an example).

The second term in the bound from Theorem 3 is roughly the larger of

$$A_1 := O\left(\frac{\Gamma_n}{\alpha n}\right) \text{ and } A_2 := \frac{R}{\alpha} \cdot \exp\left(-\Omega\left(\frac{\alpha\sqrt{n}}{\ln(1/\delta)}\right)\right)$$

(for constant $\eta$), can be compared to the lower bounds from Section 3. The lower bound from Theorem 1 is close to $A_1$ as long as $\mathbf{GES}_\rho(T, F) \approx \Gamma_n$ for $\rho = \frac{\ln 2}{2\alpha n}$. This hold for sufficiently large $n$ when $\lim_{n \to \infty} \Gamma_n = \mathbf{GES}(T, F)$. The lower bound from Theorem 2 decreases as $R \cdot \exp(-\Omega(\alpha n))$, which is a little better than $A_2$, but is otherwise qualitatively similar in terms of its dependence on the range $R^2$.

**Example 3.** If $T(F)$ is the median of $F$, and $\mathcal{F} := \{U_\gamma : \gamma \in [-R, R]\}$ is the family of uniform distributions on unit length intervals $[\gamma - 1, \gamma + 1]$ from Example 2, then $\Gamma_n = 1/2$, and the bound in Theorem 3 reduces to

$$|T(F_n) - T(F)| + O\left(\frac{1}{\alpha n}\right) + \frac{R}{\alpha} \cdot e^{-\Omega(\alpha\sqrt{n}/\ln(1/\delta))}.$$

---
[2] Appendix E shows how this discrepancy can be reduced with a stronger condition.

### 4.2. Bounding the Smooth Sensitivity

The proof of Theorem 3 (see Appendix C) is based on the following lemma, which establishes a high-probability bound on $\mathrm{SS}_\beta(T, F_n)$ under Conditions 1 and 2.

**Lemma 1.** *Assume Condition 1 and Condition 2 hold. With probability $\geq 1 - \eta$,*

$$\mathrm{SS}_\beta(T, F_n) \leq \max\left\{\frac{2\Gamma_n}{n}, R\exp\left(-\beta\left(\sqrt{\frac{n\ln(2/\eta)}{2}} - 1\right)\right)\right\}$$

*where $R$ is the quantity in Condition 1, and $\Gamma_n$ is the quantity in Condition 2.*

## 5. Differentially-Private $M$-Estimation

We now provide a procedure for constructing differentially private approximations to $M$-estimators that satisfy certain conditions. Unlike our estimators in Section 4.1, these estimators are computationally efficient; however they only apply to a more restricted class of estimators.

### 5.1. $M$-Estimators

An $M$-estimator $T_\psi(F_n)$ is given as the solution $\theta_n \in \mathbb{R}$ to the equation

$$\int \psi(x, \theta_n) dF_n(x) = 0$$

for some function $\psi \colon \mathbb{R} \times \mathbb{R} \to \mathbb{R}$. For a CDF $G$ and $\theta \in \mathbb{R}$, define

$$\Psi(G, \theta) := \int \psi(x, \theta) dG(x)$$

so $\Psi(F_n, T_\psi(F_n)) = 0$. The derivative of $\Psi$ with respect to its second argument, which is assumed to exist, is denoted by $\Psi'$. Throughout, we will assume $\psi$ satisfies the following condition.

**Condition 3** (Bounded $\psi$-range and monotonicity). *There exists a finite $K > 0$ such that the range of $\psi$ is contained in $[-K, K]$, and $\psi$ is non-decreasing in its second argument.*

Under this condition, the gross error sensitivity of $T_\psi$ at $F$ can be bounded as

$$\mathbf{GES}(T_\psi, F) = \frac{\sup_{x \in \mathbb{R}} |\psi(x, T_\psi(F))|}{|\Psi'(F, T_\psi(F))|} \leq \frac{K}{|\Psi'(F, T_\psi(F))|}. \tag{4}$$

Previous works (Chaudhuri et al., 2011) and (Rubinstein et al., 2009) have provided differentially private and computationally efficient algorithms for $M$-estimation under assumptions that are very similar to Condition 3. The algorithm in Rubinstein et al. (2009), and one of the algorithms in Chaudhuri et al. (2011) are based on the sensitivity method, while the main algorithm in Chaudhuri et al. (2011) is based on an objective perturbation method. While both algorithms are computationally efficient, both require *explicit regularization*. This is problematic in practice because determining the regularization parameter privately through differentially-private parameter-tuning requires extra data – for a more detailed discussion of this issue, see Chaudhuri et al. (2011). In contrast, our algorithm is based on the Exponential Mechanism, and does not have an explicit regularization parameter; instead we assume that $\Psi'$ is smooth, and our guarantees depend on the value of the derivative $\Psi'(F, T_\psi(F))$.

### 5.2. Exponential Mechanism for $M$-Estimation

Fix a density $\mu$ on $\mathbb{R}$, and let $\mathcal{A}_{\psi,\mu}$ be the randomized estimator whose output has probability density

$$p_{\mathcal{A}_{\psi,\mu}(F_n)}(\theta) \propto \mu(\theta) \exp\left(-\frac{n\alpha}{2K}|\Psi(F_n, \theta)|\right).$$

This estimator is derived from the exponential mechanism of McSherry & Talwar (2007), where the "cost" function is taken to be $|\Psi(F_n, \cdot)|/K$. In many $M$-estimators of interest, particularly those involving data lying in a bounded range, a prior knowledge of $K$ is reasonable.

If it is known that $T_\psi(F)$ is contained in some interval, then one can take the prior density $\mu$ to be uniform over this interval. If no such prior knowledge is available, then $\mu$ can be taken to be a density with full support on $\mathbb{R}$ such as the standard Cauchy density.

The privacy guarantee for $\mathcal{A}_{\psi,\mu}$ follows easily from known properties of the exponential mechanism (McSherry & Talwar, 2007).

**Proposition 3.** $\mathcal{A}_{\psi,\mu}$ is $(\alpha, 0)$-differentially private.

The accuracy guarantee for $\mathcal{A}_{\psi,\mu}$ relies on the following smoothness condition on $\Psi$ at $F$.

**Condition 4** (Smoothness). There exist $r_1 > 0$, $r_2 > 0$, $\Lambda_1 > 0$, and $\Lambda_2 > 0$ such that

$$|\Psi'(G, \theta) - \Psi'(F, \theta)| \leq \Lambda_1 \cdot d_{\text{GC}}(G, F) \quad \text{and}$$
$$|\Psi'(F, \theta) - \Psi'(F, T_\psi(F))| \leq \Lambda_2 \cdot |\theta - T_\psi(F)|$$

whenever $d_{\text{GC}}(G, F) \leq r_1$ and $|\theta - T_\psi(F)| \leq r_2$.

Also, for $\varepsilon > 0$ and $\eta \in (0, 1)$, define $N_{\varepsilon,\eta} := \min\{n \in \mathbb{N} : \Pr_{F_n \sim F}[|T_\psi(F_n) - T_\psi(F)| > \varepsilon] \leq \eta\}$ to be the minimum sample size such that, with probability $\geq 1 - \eta$, the non-private estimator $T_\psi(F_n)$ lies within distance $\varepsilon$ of $T_\psi(F)$.

**Theorem 4.** *Assume Condition 3 and Condition 4 hold. Let $\varepsilon_1 := \min\{r_1, |\Psi'(F, T_\psi(F))|/(6\Lambda_1)\}$, $\varepsilon_2 := \min\{r_2/2, |\Psi'(F, T_\psi(F))|/(6\Lambda_2)\}$, and $\Gamma := K/|\Psi'(F, T_\psi(F))|$. Pick any $\eta \in (0, 1)$ and $\varepsilon \in (0, \varepsilon_2)$. Suppose*

$$n \geq \max\left\{\frac{\ln(2/\eta)}{2\varepsilon_1^2},\ N_{\varepsilon_2,\eta}\right\}, \tag{5}$$

*and one of the following holds:*

1. *the range of $T_\psi$ is contained in an interval $I$ of length $R$, $\mu$ is the uniform density on $I$, and*

$$n \geq \frac{8\ln(6R/(\varepsilon\eta))}{\alpha\varepsilon} \cdot \Gamma;$$

2. *$\mu(\theta) = \frac{1}{\pi}(1 + \theta^2)^{-1}$ is the standard Cauchy density, and*

$$n \geq \frac{8}{\alpha\varepsilon} \cdot \ln\left(\frac{\pi}{\eta}\left(\frac{2(|T_\psi(F)| + \varepsilon_2)^2 + 1}{\varepsilon/3} + \frac{\varepsilon}{6}\right)\right) \cdot \Gamma.$$

*With probability at least $1 - 3\eta$, the estimator $\mathcal{A}_{\psi,\mu}$ satisfies*

$$|\mathcal{A}_{\psi,\mu}(F_n) - T_\psi(F)| \leq |T_\psi(F_n) - T_\psi(F)| + \varepsilon.$$

The proof of Theorem 4 is in Appendix D. The condition in (5) required by Theorem 4 essentially states that the sample size $n$ should be large enough for $F_n$ and $T_\psi(F_n)$ to be in the neighborhoods of $F$ and $T_\psi(F)$, respectively, where $\Psi'$ is locally Lipschitz-smooth.

It is straightforward to generalize the results to other prior densities $\mu$. Observe that in the case the range of $T_\psi$ is $[-R, R]$ for some unknown $R$, using the standard Cauchy density as $\mu$ yields a similar dependence on $R$ (via $\log |T_\psi(F)| \leq \log R$) as what is obtained when $\mu$ is uniform over $[-R, R]$. The more probability mass $\mu$ assigns around $T_\psi(F)$, the better the bounds are.

Also note that the main scaling factor of $\Gamma = K/|\Psi'(F, T_\psi(F))|$ in the sample size bound is precisely the bound on $\mathbf{GES}(T_\psi, F)$ from (4). A dependence on $\mathbf{GES}(T_\psi, F)$ is to be expected as per Theorem 1.

## 6. Conclusions

The finite sample analysis reveals a concrete connection between differential privacy and robust statistics, The main results shown here suggest using B-robustness as a criterion for designing differentially-private statistical estimators, and also highlight the obstacles that even robust estimators face when the parameter space is very large or unbounded.

While our lower bounds may seem pessimistic, they apply to estimators that succeed for a wide class of distributions. One way of avoiding our lower bounds would be by using *priors* that allow an estimator to perform well on some input distributions but not-so-well on others; a future research direction is to investigate how this can help design better differentially private estimators.

**Acknowledgements.** KC would like to thank NIH U54 HL108460 for research support.